# A Comprehensive Analysis of Real-World Image Captioning and Scene Identification

Sai Suprabhanu Nallapaneni, Subrahmanyam Konakanchi
nallapaneni.s@ufl.edu, konakanchi.s@ufl.edu

**Abstract:**

Image captioning is a computer vision task that involves generating natural language descriptions for images. This method has numerous applications in various domains, including image retrieval systems, medicine, and various industries. However, while there has been significant research in image captioning, most studies have focused on high quality images or controlled environments, without exploring the challenges of real-world image captioning. Real-world image captioning involves complex and dynamic environments with numerous points of attention, with images which are often very poor in quality, making it a challenging task, even for humans. This paper evaluates the performance of various models that are built on top of different encoding mechanisms, language decoders and training procedures using a newly created real-world dataset that consists of over 800+ images of over 65 different scene classes, built using MIT Indoor scenes dataset. This dataset is captioned using the IC3 approach that generates more descriptive captions by summarizing the details that are covered by standard image captioning models from unique view-points of the image.

## 1. Introduction:

Image captioning can be thought of as translating an image's features and attributes into a meaningful sentence. This method has numerous applications in various domains, including image retrieval systems, medicine, and various industries. In image retrieval systems, image captioning can be used to provide more accurate and relevant search results by allowing users to search for images using text-based queries. In the medical field, image captioning can be used to automatically generate captions for medical images, aiding in diagnosis and treatment planning. In industrial sectors, image captioning can be used for automated quality control and visual inspection, as well as in autonomous systems such as self-driving cars.

Image captioning is a vision-language modeling task that has seen remarkable progress over the years, thanks to advancements in computer vision and natural language processing techniques. The earliest image captioning models were based on the visual encoding of images, which were then mapped to natural language descriptions using simple neural networks[16]. As the field progressed, language models such as Recurrent Neural Networks (RNNs)[17] and Long Short-Term Memory (LSTM) networks[1] were introduced to generate more complex and coherent sentences. To further improve the performance and generate more coherent captions, an attention mechanism[2,3,20] was introduced. More recent advancements in transformer models with self attention mechanisms[7], BERT[18] and GPT-3[15], have revolutionized image captioning, enabling models to generate more accurate and contextually relevant captions by learning from vast amounts of textual data. In addition to architecture, training strategies such as reinforcement learning[4,5] and vision-language pre-training(VLP)[18,19] have also played an important role in improving image captioning performance.

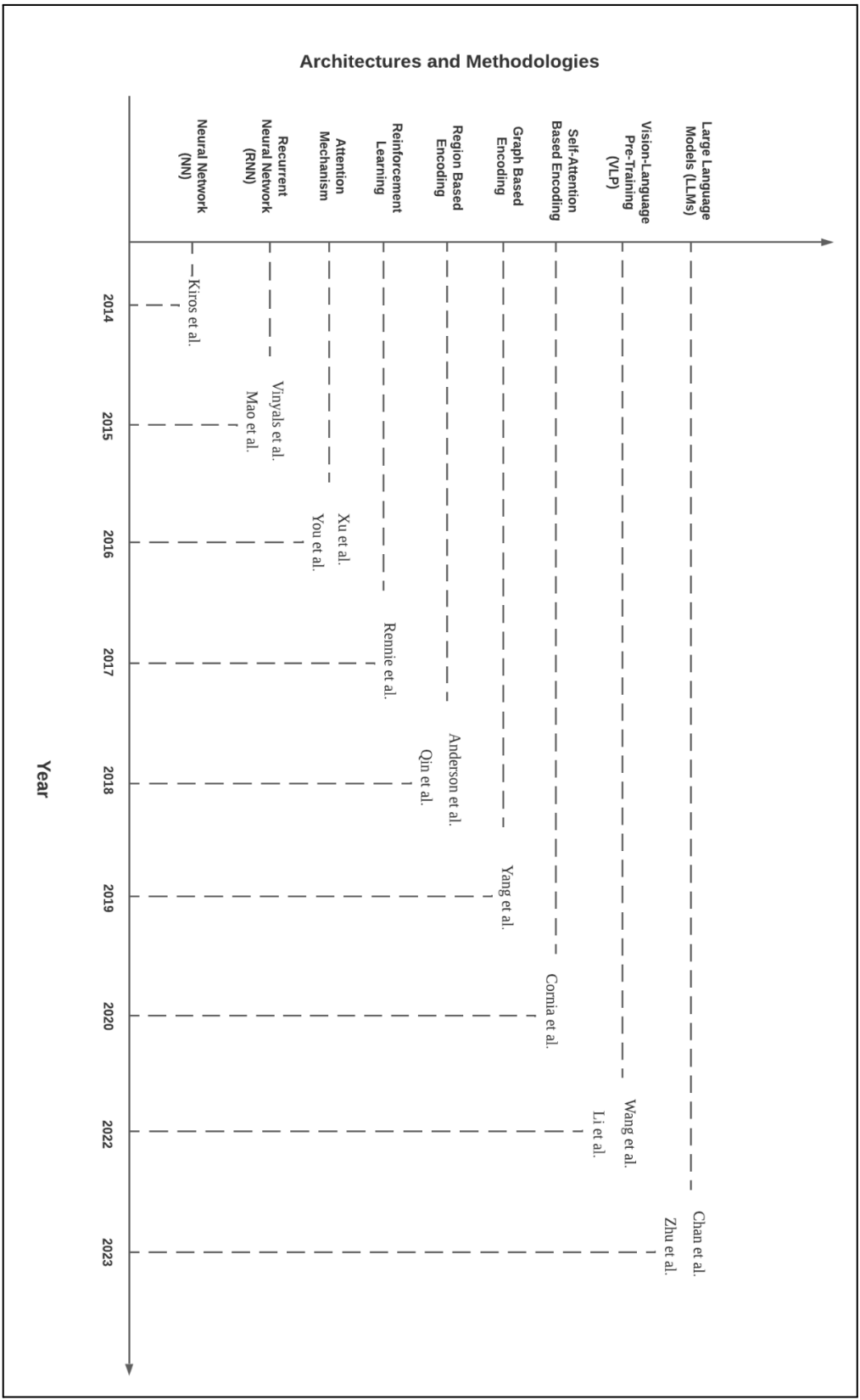

**Figure:1** Illustration of the timeline depicting the development of different architectures and methodologies in the field of image captioning.

Although these models have achieved state-of-the-art performance in image captioning tasks, most studies have focused on simple images or controlled environments, without exploring the challenges of real-world image captioning or understanding the real-world scenes. [Fig 2, 3] illustrate the factors that contribute to the challenges encountered in real-world scene identification. In this survey paper, we evaluate the performance of various models built on different architectures and methodologies, focusing on those that achieved comparable results on evaluation metrics. Our aim is to identify which architecture and methodology performs well in the challenging task of real-world image captioning and scene identification. To assess the models in real-world scenarios, we will develop a new dataset consisting of over 800+ indoor and outdoor scene images, built using MIT Indoor Scenes dataset. To ensure high-quality captions for the images, we will generate the captions using a model that uses the IC3 approach proposed by [14], instead of manual generation, as this model has yielded good results in capturing most of the details in a given scene, and this procedure will be discussed in detail in the following sections.

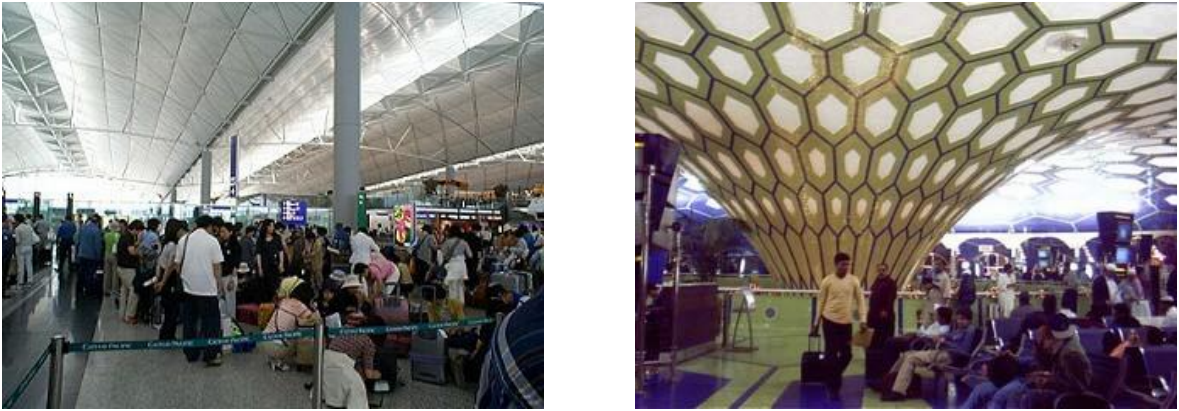

**Figure 2, 3**: These images contain - a large group of people standing and sitting in an airport, possibly waiting in line for their plane. Captioning these images using a model becomes challenging as the visual features that strongly support the word "airport" are scarce. It is not only difficult for machines, but even humans can face similar challenges when interpreting such images.

## 1.1. Contributions:

The primary objective of our work is to present a comprehensive analysis of various models using different architectures and methodologies for real-world image captioning and scene identification tasks. While several notable works have surveyed image captioning techniques, architectures, and evaluation metrics [10,11,12], we will focus on evaluating models on real-world scenes dataset. To facilitate our analysis, we developed a new dataset consisting of over 65 distinct scene classes using the MIT indoor scenes dataset. To generate captions for evaluation, we used IC3-based image description instead of manual generation since this model has demonstrated good results on new human evaluation metrics based on "Helpfulness" and "Correctness". Along with the dataset, we also introduced a novel evaluation metric named Scene Identification Score (SIS) specifically for scene identification. By comparing the performance of various models on our real-world dataset, we aim to identify the most effective architectures and methodologies for image captioning and scene identification in real-world scenarios.

**TABLE 1**: The study provides an analysis of the performance metrics(mode values) of three different architectures utilized by five different reference models. This analysis demonstrates how well these architectures perform when compared to the Unified VLP model (given in the last row), which incorporates Large-Scale Vision-Language Pre-Training.

| Approach | BLEU-4 | METEOR | CIDEr |
|---|---|---|---|
| Region Based + RNN/LSTM [5, 21, 24, 25, 26] | 38 | 28 | 126 |
| Graph Based + RNN/LSTM [6, 20, 27, 28, 29 ] | 38 | 28 | 128 |
| Self-Attention Based + RNN/LSTM [7, 30, 31, 32, 33 ] | 39 | 29 | 129 |
| BERT + VLP [Unified VLP] | 39.5 | 29.3 | 129.3 |

## 2. Literature Survey:

We have conducted an extensive examination of different studies and architectures in the field of image captioning. We have explored the advantages and limitations of each approach, starting from the initial neural network-based captioning model. We then moved on to attention-based models, followed by transformer-based models. Next, we studied vision language pretrainers and finally Generative Pretrained Transformer-based models. This thorough survey of these studies will greatly assist us in analyzing and understanding real-world image captioning and scene identification.

**2.1.** Show and Tell [2015]

"Show and Tell" by Vinyals et al. is one of the earliest and state-of-the-art works in the domain of image captioning. It introduced the implementation of a Recurrent Neural Network (RNN) as a language model. This model outperformed Kiros et al., who proposed the first Neural Network (Feedforward Network), and Mao et al., who introduced the first Recurrent NN, in terms of evaluation metrics. This was achieved by directly providing the visual input to the RNN and making changes to the encoding approach. However, employing RNNs as decoders in image captioning becomes an apparent idea when considering it as a language translation task, where the goal is to translate an image into text rather than translating between different spoken languages. In contrast to the encoding process in language translation tasks, where RNNs encode words into a hidden state, image captioning leverages CNNs as encoders to effectively capture visual features. The model utilizes Global Encoding with CNN, RNN/LSTM as the decoder and employs the Cross Entropy Loss training strategy.

## 2.2 Show, Attend and Tell [2015]

Attention plays a crucial role in our ability to focus on important aspects while performing any task. This paper by Xu et al. extended the "Show and Tell" [1] model by introducing a visual attention mechanism to improve the alignment between the visual features and the generated words. The attention mechanism allows the model to selectively focus on different regions of the image at each time step during caption generation. By attending to the relevant regions, the model can generate more accurate and contextually appropriate captions. This study introduces two variants: soft attention, which enables the model to focus on different parts of the input sequence and attend to multiple elements simultaneously, and hard attention, which selects only a subset of elements in a given sequence. The language model and training strategy employed in this approach are broadly similar to those of [1] except for the encoding strategy. Unlike [1] which utilizes global encoding, this model employs a grid-based encoding technique that leads to improved feature extraction.

## 2.3. Image Captioning with Semantic Attention [2016]

This study by You et al. builds upon the concept of visual attention, similar to the approach presented in [2]. However, notable modifications in the implementation and architecture of this model enabled it to surpass various state-of-the-art models that were previously on par. Unlike the fixed resolution spatial attention modeling in [2], this model allows for the utilization of attention concepts from any location and resolution within the image, even if they are not visually present. Additionally, the model incorporates a feedback process that combines top-down and bottom-up features. Instead of using pre-trained features at specific spatial locations like in [2], this model utilizes word features corresponding to detected visual concepts. Similar to [1], this model also utilizes Global Encoding with CNN, RNN/LSTM as the decoder and employs the Cross Entropy Loss training strategy.

## 2.4. Self-critical Sequence Training for Image Captioning [2017]

SCST by Rennie et al. is one of the notable works in the field of image captioning as it has provided many insights to a large number of state of the art works of the future. This model learns from its own predictions during testing to evaluate and improve its performance. It doesn't rely on external evaluations but uses its own judgment to assess how well it is doing and make adjustments accordingly. This means that only the samples from the model that perform better than the current system during testing are given importance, while the samples that are not as good are given less attention or suppressed. Image features are encoded using a deep CNN, similar to the approach described in [2], with a few modifications in the architecture of the attention model (Att2in). In this model, the image feature derived from attention is only inputted to the cell node of the LSTM. The encoding strategy employed is grid-based, with an RNN/LSTM decoder, and training is conducted using cross-entropy loss and reinforcement learning techniques.

**2.5.** Bottom-Up and Top-Down Attention for Image Captioning [2018]

In their study, Anderson et al. propose a combined bottom-up and top-down attention mechanism, inspired by the work in [3]. Unlike previous methods that used global or grid-based encoding, this study introduces a region-based encoding approach based on Faster R-CNN which consists of a Region Proposal Network (RPN) in the first stage and a RoI (Region of Interest) pooling layer in the second stage. This allows for easy extraction of objects and salient regions in the image. The captioning model they propose, while similar to previous approaches [2,4], incorporates specific architectural changes and achieves state-of-the-art performance even without bottom-up attention. They also employ a reinforcement learning approach similar to [4], but with the addition of sampling distribution restrictions for a reduced training time. By combining the new region-based encoding technique, well-structured top-down attention, language LSTM models, and efficient reinforcement learning as the training strategy, this model achieves breakthroughs and state-of-the-art performance in image captioning.

**2.6.** Auto-Encoding Scene Graphs for Image Captioning [2019]

When we humans imagine a boat, we naturally assume that it is on water, even if the water is not visible. This kind of assumption is called the inductive bias. To generate more detailed and high-quality captions that mimic human understanding, Yang et al. proposed a method called Scene Graph Auto-Encoder (SGAE). This method incorporates the inductive bias into the encoder-decoder pipeline by using a shared dictionary. The pipeline includes a graph encoder based on Graph Convolutional Networks (GCN) to encode the scene graph, and a visual encoder based on Region Proposal Networks (RPN). The captioning model also includes a language model similar to [5] and is trained using a reinforcement learning strategy similar to [4]. Indeed, most of this study is based on the previous works [4,5], including the other mentioned studies, ultimately resulting in the achievement of state-of-the-art performance by this model. In summary, this model utilizes graph and region-based encoding, an RNN/LSTM decoder, and is trained using reinforcement learning.

**2.7.** Meshed-Memory Transformer for Image Captioning [2020]

After the groundbreaking study "Attention is all you need" by Vaswani et al. in 2017, it has become evident that Transformers outperform every other architecture in generation and translation tasks. In contrast to prior methods, Cornia et al. present a fully-attentive model that draws inspiration from the work of Vaswani et al. This model utilizes self-attention and eliminates the necessity for recurrence and convolution layers. Despite this difference, the model can still be conceptualized as an encoder-decoder pipeline. The encoder consists of stacked encoding layers, each incorporating an attention mechanism, feed-forward layer, and two memory vectors. These layers collectively refine the relationships between image regions by leveraging a priori knowledge encoded in the memory vectors, similar to the notion of inductive bias [6]. In contrast to recurrent networks, which struggle with long-range dependencies, self-attention computes attention scores between all pairs of elements in the sequence, weighting their contributions to the representation of other elements. Each encoder layer is connected to a decoder layer, forming a mesh-like structure with learnable connectivity. Thus, due to its mesh-like structure and the utilization of memory vectors to refine relationships between image regions, this model is named as Meshed Memory Transformer.

**2.8.** BLIP: Bootstrapping Language-Image Pre-training [2022]

The paper "BLIP: Bootstrapping Language-Image Pretraining" introduces a novel approach that leverages large-scale pretraining to capture billions of parameters while training on vast image datasets. By incorporating both text and image data, the authors address the limitations of text-only pre training methods. Their bootstrapping framework integrates image features into the process and explores the effectiveness of combining textual and visual information in language model (LM) pretraining. The study demonstrates significant improvements in downstream tasks like image captioning and visual question answering. The authors propose a contrastive learning objective to align image and text embeddings and promote the learning of semantically meaningful representations. They also investigate strategies like multimodal transformers and cross-modal distillation to incorporate visual data. The results highlight the advantages of jointly leveraging text and image data, surpassing the performance of traditional text-only pretraining. The paper contributes to multimodal learning research and showcases the potential of integrating visual and textual information in LM pretraining, enabling better understanding and generation of multimodal content.

**2.9.** Image Captioning by Committee Consensus [2023]

This model proposed by Chan et al. combines the state-of-the-art models OFA or BLIP, and the GPT3. In simpler terms, this method involves selecting an arbitrary number of captions, which are generated by advanced image captioning engines like OFA using a technique called temperature-based sampling. These captions are then summarized using a powerful generative engine such as the GPT-3's Davinci-v3. The authors of this study put considerable effort into crafting a well-designed prompt, which enabled the language model to generate highly detailed summaries that even surpassed the state-of-the-art OFA model. However, evaluating the performance of this model using standard metrics proved to be inadequate due to the high level of detail in the generated captions. Instead, the model's effectiveness was assessed using new human evaluation metrics focused on "Helpfulness" and "Correctness". Remarkably, this method resulted in captions that are significantly more descriptive and informative compared to those generated by individual models.

**2.10.** Chat Captioner [2023]

Zhu et al. introduced a novel automatic questioning system called ChatCaptioner for image captioning. Building upon the use of Large-Scale Pre-trained models and Large Language Models as seen in previous works like [9], this approach combines the strengths of BLIP-2, a robust image captioning and visual question answering (VQA) model, with ChatGPT, a powerful Large Language Model. ChatGPT interacts with BLIP-2 by posing questions to its VQA engine, extracting additional image details. By leveraging visual question answering, automatic question generation, and conversation summarization, this method produces more detailed captions that cover multiple aspects of the image. While traditional evaluation metrics like BLEU score show lower performance compared to models like BLIP-2, the quality of the generated captions is significantly improved. To evaluate their approach, the authors employed Human Votes for assessing image information and conducted Correctness Analysis, demonstrating that this model outperforms other models in both aspects.

## 3. Methodology:

In this study, we utilize different models, each constructed with distinct architectures. These architectures are based on visual encoding, language models, and training strategies. The first approach employs region-based encoding with RNN/LSTM as the language model and is trained using cross-entropy loss and reinforcement learning strategies (Sec 2.1). The second approach utilizes graph-based encoding with RNN/LSTM as the language model and is also trained using cross-entropy loss and reinforcement learning (Sec 2.2). The third approach adopts self-attention-based encoding with a transformer as the language model and is trained using cross-entropy loss and reinforcement learning strategies (Sec 2.3). It has been observed that despite the existence of multiple models based on the above three approaches, they perform comparably (Table-1). Despite the differences in their encoding strategies, these three methods share a common decoder and training strategies. This study also provides insights into the effectiveness of various encoding strategies for real-world scene identification. Table-1 also demonstrates the performance of these architectures compared to the Unified VLP model (shown in the last row), which incorporates Large-Scale Vision-Language Pre-Training. Other encoding strategies such as Global and Grid-based encoding were excluded due to significant differences in performance metrics, which would introduce unfair comparisons.

Typically, an image captioning pipeline consists of an encoder and a decoder [Fig 4]. The encoder utilizes a CNN to extract the features from the input image, while the decoder generates captions word by word using the encoded features through an RNN. In addition, a suitable training procedure is employed to train these components effectively. In our analysis, we will provide a comprehensive overview of three models, focusing on their encoding, decoding, and training strategies.

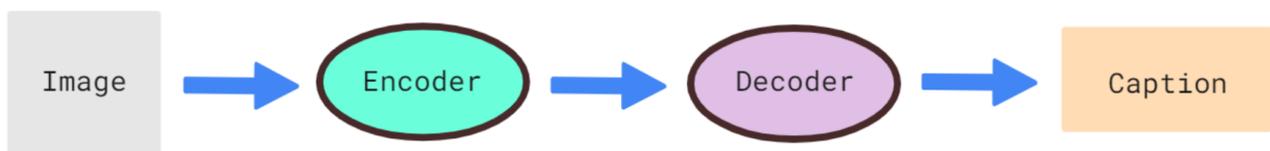

**Fig 4**: A simple encoder-decoder pipeline for image captioning

**Visual Encoding or Encoders**

Encoding the visual features of an image is a crucial and fundamental step in generating detailed and precise captions. This becomes particularly significant when dealing with real-world images and scenes that encompass a multitude of intricate features and details. The encoder plays a pivotal role in capturing these details from various points of attention. The illustration in [Fig 5, 6, 7] demonstrates the functioning of region-based, graph-based, and self-attention-based encoding mechanisms respectively in an image.

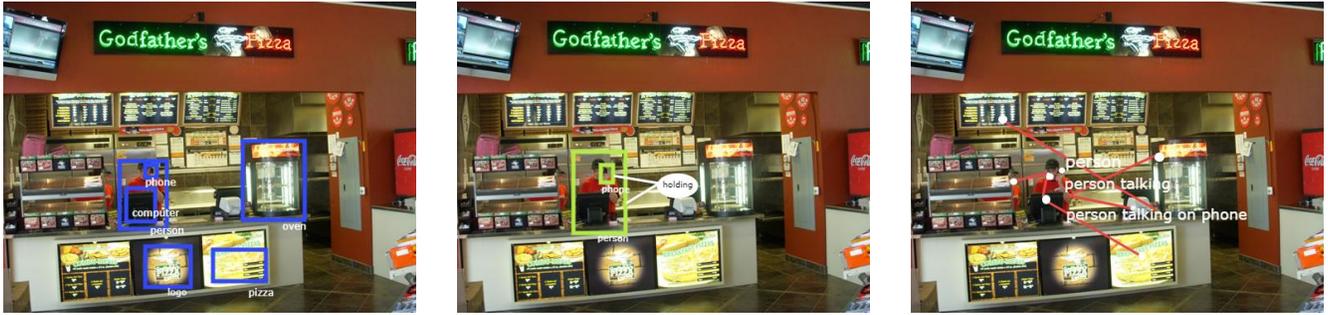

**Fig 5, 6, 7:** depict the encoding mechanisms used in this study. Figure 5 illustrates the visual regions used for region-based encoding, Figure 6 represents the graphical representations used for graph-based encoding, and Figure 7 showcases the point-to-point relationships utilized in the self-attention based encoding mechanism.

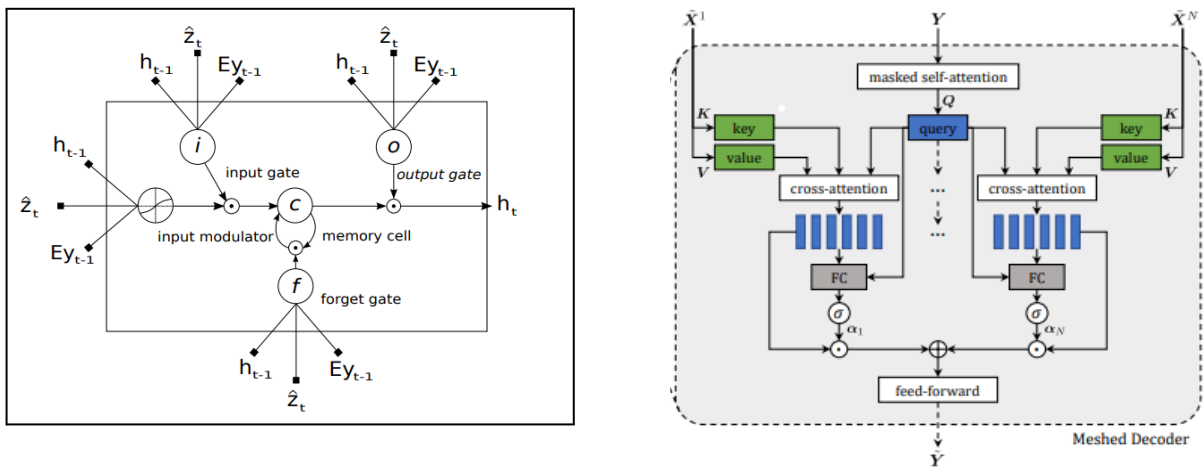

**Fig 8, 9:** An LSTM cell in a Recurrent Neural Network [2], Self-attention based transformer in a decoding layer [7] respectively

**Language Models or Decoders**

The main aim of a decoder or language model is to transform the encoded visual features into meaningful and coherent descriptions by estimating the probabilities of a given sequence of words appearing in a sentence. In our analysis, we explore Recurrent Neural Networks (RNNs) with Long Short-Term Memory (LSTM)[Fig 8] and Transformers[Fig 9].

**Training Procedures**

The development of a robust captioning pipeline relies heavily on an appropriate training procedure. In the domain of image captioning, two commonly employed training methods are Cross-Entropy Loss and Reinforcement Learning.

### 3.1. Region Based Encoding - RNN/LSTM - Reinforced, Cross-Entropy Loss Training

In their study, Anderson et al. propose a combined bottom-up and top-down attention mechanism, inspired by the work in [3]. Unlike previous methods that used global or grid-based encoding, this study introduces a region-based encoding approach based on Faster R-CNN which consists of a Region Proposal Network (RPN) in the first stage and a RoI (Region of Interest) pooling layer in the second stage. This allows for easy extraction of objects and salient regions in the image. The captioning model they propose, while similar to previous approaches [2,4], incorporates specific architectural changes and achieves state-of-the-art performance even without bottom-up attention.

They also employ a reinforcement learning approach similar to [4], but with the addition of sampling distribution restrictions for a reduced training time. By combining the new region-based encoding technique, well-structured top-down attention, language LSTM models, and efficient reinforcement learning as the training strategy, this model achieves breakthroughs and state-of-the-art performance in image captioning. However, for generating captions, we employed [21] as our chosen method, which is derived from [3], primarily because it offers advantages in terms of reduced resource usage and time requirements compared to the original [3] approach.

### 3.2. Graph Based Encoding - RNN/LSTM - Reinforced, Cross-Entropy Loss Training

When we humans imagine a boat, we naturally assume that it is on water, even if the water is not visible. This kind of assumption is called the inductive bias. To generate more detailed and high-quality captions that mimic human understanding, Yang et al. proposed a method called Scene Graph Auto-Encoder (SGAE). This method incorporates the inductive bias into the encoder-decoder pipeline by using a shared dictionary. The pipeline includes a graph encoder based on Graph Convolutional Networks (GCN) to encode the scene graph, and a visual encoder based on Region Proposal Networks (RPN).

The captioning model also includes a language model similar to [5] and is trained using a reinforcement learning strategy similar to [4]. Indeed, most of this study is based on the previous works [4,5], including the other mentioned studies, ultimately resulting in the achievement of state-of-the-art performance by this model. In summary, this model utilizes graph and region-based encoding, an RNN/LSTM decoder, and is trained using reinforcement learning.

### 3.3. Self-Attention Based Encoding - Transformer - Reinforced, Cross-Entropy Loss Training

After the groundbreaking study "Attention is all you need" by Vaswani et al. in 2017, it has become evident that Transformers outperform every other architecture in generation and translation tasks. In contrast to prior methods, Cornia et al. present a fully-attentive model that draws inspiration from the work of Vaswani et al. This model utilizes self-attention and eliminates the necessity for recurrence and convolution layers. Despite this difference, the model can still be conceptualized as an encoder-decoder pipeline. The encoder consists of stacked encoding layers, each incorporating an attention mechanism, feed-forward layer, and two memory vectors. These layers collectively refine the relationships between image regions by leveraging a priori knowledge encoded in the memory vectors, similar to the notion of inductive bias [6].

In contrast to recurrent networks, which struggle with long-range dependencies, self-attention computes attention scores between all pairs of elements in the sequence, weighting their contributions to the representation of other elements. Each encoder layer is connected to a decoder layer, forming a mesh-like structure with learnable connectivity. Thus, due to its mesh-like structure and the utilization of memory vectors to refine relationships between image regions, this model is named as Meshed Memory Transformer.

**4. Dataset:**

The majority of studies have primarily focused on training, testing, and validating their models using three main datasets: MS COCO, Flickr30k, and Flickr8k [10,11,12]. These datasets were developed within controlled environments, consisting of high-quality images and accompanying captions. However, real-world scene images often exhibit distortions and lower quality. In our study, it is crucial to validate these models using a more specific dataset that represents real-world scenes rather than a generic one. While there is currently no dataset exclusively dedicated to real-world scenes, we have created our own dataset for this purpose. Instead of manually captioning the images in the dataset, which would practically be not possible given our limited resources and time, we adopted a different approach by utilizing a sophisticated model proposed by [14] that combines the state-of-the-art model OFA, and the GPT3. In simpler terms, this method involves selecting an arbitrary number of captions, which are generated by advanced image captioning engines like OFA using a technique called temperature-based sampling. These captions are then summarized using a powerful language model such as GPT3. This method resulted in captions that are significantly more descriptive and informative compared to those generated by individual models. [Fig 10, 11, 12] displays the images alongside the corresponding generated captions.

Our new validation dataset contains nearly 800 different images belonging to about 65 different scene categories. We used a 10 GB T4 GPU in addition to a 12 GB CPU for the computation. For each image, we generated two captions, and the entire process for the entire dataset took approximately 15 hours. Out of every 100 captions, we randomly selected 10 for manual qualitative analysis. The majority of these captions correctly described the scene and included all the identifiable details.

## 5. Evaluation Metrics:

In our evaluation, we assess models 3.1, 3.2, and 3.3 based on their performance in real-world image captioning and scene identification, taking into account the specific objective at hand. While it is true that image captioning and scene identification are interconnected tasks, as both rely on the quality of the encoder and decoder architecture, as well as the training procedure employed, it is the underlying intuition, inductive bias, or prior knowledge that enables the model to make conclusions about a particular scene based on extracted features, details, attributes, or objects. Therefore, we believe it is necessary to evaluate these two tasks separately in order to gain a comprehensive understanding of the models' capabilities.

For evaluating the quality of image captioning, various metrics such as BLEU, METEOR, ROUGE, and CIDEr are commonly used. In our analysis, we have opted to utilize the BLEU metric. However, when it comes to scene identification, there is currently no established metric available. To address this gap, we have introduced a novel metric called Scene Identification Score (SIS).

### 5.1. BLEU (Bilingual Evaluation Understudy):

The BLUE metric, short for Bilingual Evaluation Understudy, is a widely used measure in natural language processing (NLP) to evaluate the quality of machine-generated translations. The primary goal of the BLUE metric is to assess the similarity between the machine-generated translation and the human references. It takes into account both the precision of matching n-grams and the brevity penalty to penalize excessively short translations. The higher the BLUE score, which ranges from 0 to 1, the closer the machine-generated translation is to the human references.

### 5.2 . SIS (Scene Identification Score):

The concept behind SIS is straightforward: for each scene class, we measure the percentage of captions generated by a model that accurately identifies the corresponding scene. The SIS score for a model is then calculated as the average of these percentages across all scene classes. This metric provides a means to assess the performance of models in scene identification. To compute the SIS score, we leveraged the OpenAI's Davinci engine API, employing a meticulously crafted prompt design.

## 6. Results and Discussion:

We have employed the SIS metric in conjunction with the LBPF image captioning model to assess its effectiveness in analyzing and identifying the overall scene depicted in an image. To conduct our evaluation, we selected approximately 36 categories of images from the MIT indoor scenes dataset. After generating captions for these images using LBPF, we compared the captions with the corresponding image categories to determine if the model successfully recognized the scene. Unfortunately, the results of this comparison were not satisfactory. The model exhibited poor performance on the majority of the tested images, struggling particularly with images containing complex features such as "movie theater," "cloister," and "museum." It appears that the model's shortcomings in real-world scene identification and captioning stem from a lack of inductive bias [6] or prior knowledge [7].

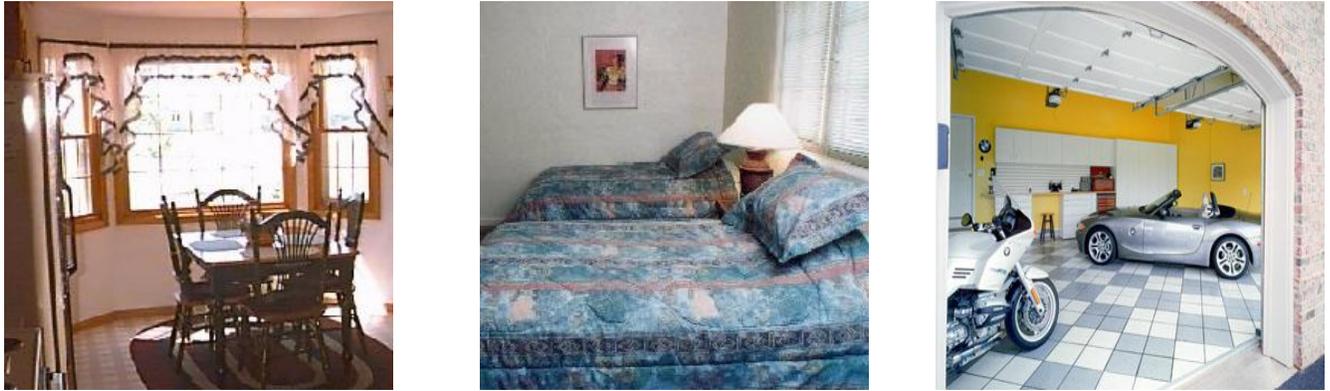

**Fig 10 (left): a.)** `A dining room with a wooden table and chairs, a refrigerator, and two windows.` **b.)** `A dining room with a wooden table and chairs in front of a window, and possibly two or three windows in the room.`

**Fig 11 (center): a.)** `A bedroom with two beds, a lamp, and possibly a window.` **b.)** `A hotel room with two twin beds and a lamp near a window.`

**Fig 12 (right): a.)** `A motorcycle parked in a garage next to a convertible sports car.` **b.)** `A garage containing a motorcycle and a convertible sports car parked next to each other.`

`Table-2: Results of image scene detection are mentioned in the following:`

| Image_Category | Number of Images | Scene Detection Count | Percentage |
|---|---|---|---|
| Airport | 12 | 1 | 8.4 |
| Auditorium | 12 | 0 | 0 |
| Bakery | 8 | 1 | 12.5 |
| Bar | 24 | 3 | 12.5 |
| Bathroom | 11 | 9 | 81.9 |
| Bedroom | 11 | 5 | 45.5 |
| Bookstore | 12 | 5 | 41.7 |
| Casino | 15 | 0 | 0 |
| Church | 11 | 2 | 18.2 |
| Classroom | 13 | 0 | 0 |
| Cloister | 16 | 0 | 0 |

| Closet | 16 | 0 | 0 |
| --- | --- | --- | --- |
| ClothingStore | 5 | 0 | 0 |
| ComputerRoom | 10 | 3 | 30 |
| ConcertHall | 10 | 0 | 0 |
| Corridor | 12 | 0 | 0 |
| Deli | 8 | 4 | 50 |
| DentalOffice | 6 | 0 | 0 |
| DiningRoom | 16 | 0 | 0 |
| Elevator | 7 | 0 | 0 |
| FastFood Restaurant | 13 | 2 | 15.4 |
| Florist | 8 | 0 | 0 |
| Grocery Store | 17 | 0 | 0 |
| Gym | 16 | 0 | 0 |
| HairSalon | 9 | 0 | 0 |
| HospitalRoom | 15 | 0 | 0 |
| Bus | 14 | 8 | 57.2 |
| Subway | 16 | 0 | 0 |
| Kindergarten | 7 | 1 | 14.3 |
| Kitchen | 19 | 14 | 73.7 |
| laundromat | 12 | 0 | 0 |
| Library | 9 | 2 | 22.3 |
| Mall | 10 | 0 | 0 |
| Movie Theatre | 13 | 0 | 0 |
| Museum | 6 | 0 | 0 |
| SwimmingPool | 12 | 11 | 91.7 |

## 7. Conclusion and Future Work:

We have presented a comprehensive timeline and literature review of different models used in image captioning, covering various architectures and methodologies. Starting from the initial Feed Forward Neural Network-based model to the latest GPT-based models, we have highlighted the advancements and improvements made over the years. Additionally, we have identified and discussed the challenges faced in real-world image captioning, as well as conducted a detailed analysis of real-world scene identification using three different architectures. Furthermore, we have introduced a novel evaluation metric called SIS (Scene Identification Score) specifically for real-world scene identification.

Moving forward, our future work will concentrate on the application of Large Scale Pretraining to a generic CNN/RNN-based encoder-decoder pipeline, which remains largely unexplored despite being mentioned in several survey papers. We plan to explore training existing models on state-of-the-art image networks, incorporating new training strategies and leveraging advanced technologies. Moreover, considering the recent advancements in large language models like GPT-4, we are interested in exploring the emerging possibilities in the image captioning domain presented by these models.

## 8. Acknowledgements:

We express our sincere gratitude to Prof. Radha Guha for her invaluable guidance and mentorship throughout our study. We are grateful to SRM University AP for generously providing us with the necessary computational resources. Additionally, we would like to acknowledge the support of the OpenAI team for their provision of the partial free tire API, which significantly reduced our computational costs. We extend our thanks to the open source projects and the supportive community that played a crucial role in our research journey.

# References

1. Vinyals, O., Toshev, A., Bengio, S., & Erhan, D. (2015). Show and tell: A neural image caption generator. In *Proceedings of the IEEE conference on computer vision and pattern recognition* (pp. 3156-3164).

2. Xu, K., Ba, J., Kiros, R., Cho, K., Courville, A., Salakhudinov, R., ... & Bengio, Y. (2015, June). Show, attend and tell: Neural image caption generation with visual attention. In *International conference on machine learning* (pp. 2048-2057). PMLR.

3. You, Q., Jin, H., Wang, Z., Fang, C., & Luo, J. (2016). Image captioning with semantic attention. In *Proceedings of the IEEE conference on computer vision and pattern recognition* (pp. 4651-4659).

4. Rennie, S. J., Marcheret, E., Mroueh, Y., Ross, J., & Goel, V. (2017). Self-critical sequence training for image captioning. In *Proceedings of the IEEE conference on computer vision and pattern recognition* (pp. 7008-7024).

5. Anderson, P., He, X., Buehler, C., Teney, D., Johnson, M., Gould, S., & Zhang, L. (2018). Bottom-up and top-down attention for image captioning and visual question answering. In *Proceedings of the IEEE conference on computer vision and pattern recognition* (pp. 6077-6086).

6. Yang, X., Tang, K., Zhang, H., & Cai, J. (2019). Auto-encoding scene graphs for image captioning. In *Proceedings of the IEEE/CVF conference on computer vision and pattern recognition* (pp. 10685-10694).

7. Cornia, M., Stefanini, M., Baraldi, L., & Cucchiara, R. (2020). Meshed-memory transformer for image captioning. In *Proceedings of the IEEE/CVF conference on computer vision and pattern recognition* (pp. 10578-10587).

8. Li, J., Li, D., Xiong, C., & Hoi, S. (2022, June). Blip: Bootstrapping language-image pre-training for unified vision-language understanding and generation. In *International Conference on Machine Learning* (pp. 12888-12900). PMLR.

9. Wang, P., Yang, A., Men, R., Lin, J., Bai, S., Li, Z., ... & Yang, H. (2022, June). Ofa: Unifying architectures, tasks, and modalities through a simple sequence-to-sequence learning framework. In *International Conference on Machine Learning* (pp. 23318-23340). PMLR.

10. Hossain, M. Z., Sohel, F., Shiratuddin, M. F., & Laga, H. (2019). A comprehensive survey of deep learning for image captioning. *ACM Computing Surveys (CsUR)*, *51*(6), 1-36.

11. Stefanini, M., Cornia, M., Baraldi, L., Cascianelli, S., Fiameni, G., & Cucchiara, R. (2022). From show to tell: A survey on deep learning-based image captioning. IEEE transactions on pattern analysis and machine intelligence, 45(1), 539-559.

12. Xu, Y., Li, L., Xu, H., Huang, S., Huang, F., & Cai, J. (2022). Image Captioning In the Transformer Age. *arXiv preprint arXiv:2204.07374*.

13. Vaswani, A., Shazeer, N., Parmar, N., Uszkoreit, J., Jones, L., Gomez, A. N., ... & Polosukhin, I. (2017). Attention is all you need. *Advances in neural information processing systems*, *30*.

14. Chan, D. M., Myers, A., Vijayanarasimhan, S., Ross, D. A., & Canny, J. (2023). $IC^3$: Image Captioning by Committee Consensus. *arXiv preprint arXiv:2302.01328*.

15. Zhu, D., Chen, J., Haydarov, K., Shen, X., Zhang, W., & Elhoseiny, M. (2023). Chatgpt asks, blip-2 answers: Automatic questioning towards enriched visual descriptions. *arXiv preprint arXiv:2303.06594*.

16. Kiros, R., Salakhutdinov, R., & Zemel, R. (2014, June). Multimodal neural language models. In *International conference on machine learning* (pp. 595-603). PMLR.

17. Mao, J., Xu, W., Yang, Y., Wang, J., & Yuille, A. L. (2014). Explain images with multimodal recurrent neural networks. *arXiv preprint arXiv:1410.1090*.

18. Zhang, P., Li, X., Hu, X., Yang, J., Zhang, L., Wang, L., ... & Gao, J. (2021). Vinvl: Revisiting visual representations in vision-language models. In *Proceedings of the IEEE/CVF Conference on Computer Vision and Pattern Recognition* (pp. 5579-5588).

19. Zhou, L., Palangi, H., Zhang, L., Hu, H., Corso, J., & Gao, J. (2020, April). Unified vision-language pre-training for image captioning and vqa. In *Proceedings of the AAAI conference on artificial intelligence* (Vol. 34, No. 07, pp. 13041-13049).

20. Yao, T., Pan, Y., Li, Y., & Mei, T. (2019). Hierarchy parsing for image captioning. In *Proceedings of the IEEE/CVF international conference on computer vision* (pp. 2621-2629).

21. Qin, Y., Du, J., Zhang, Y., & Lu, H. (2019). Look back and predict forward in image captioning. In Proceedings of the IEEE/CVF conference on computer vision and pattern recognition (pp. 8367-8375).

22. Zhu, D., Chen, J., Haydarov, K., Shen, X., Zhang, W., & Elhoseiny, M. (2023). Chatgpt asks, blip-2 answers: Automatic questioning towards enriched visual descriptions. arXiv preprint arXiv:2303.06594.

23. Li, J., Li, D., Savarese, S., & Hoi, S. (2023). Blip-2: Bootstrapping language-image pre-training with frozen image encoders and large language models. arXiv preprint arXiv:2301.12597.

24. L. Huang, W. Wang, Y. Xia, and J. Chen, "Adaptively Aligned Image Captioning via Adaptive Attention Time," in NeurIPS, 2019.

25. L. Wang, Z. Bai, Y. Zhang, and H. Lu, "Show, Recall, and Tell: Image Captioning with Recall Mechanism," in AAAI, 2020.

26. Z.-J. Zha, D. Liu, H. Zhang, Y. Zhang, and F. Wu, "Context-aware visual policy network for fine-grained image captioning," IEEE Trans. PAMI, 2019.

27. T. Yao, Y. Pan, Y. Li, and T. Mei, "Exploring Visual Relationship for Image Captioning," in ECCV, 2018.

28. L. Guo, J. Liu, J. Tang, J. Li, W. Luo, and H. Lu, "Aligning linguistic words and visual semantic units for image captioning," in ACM Multimedia, 2019

29. Z. Shi, X. Zhou, X. Qiu, and X. Zhu, "Improving Image Captioning with Better Use of Captions," in ACL, 2020.

30. X. Yang, H. Zhang, and J. Cai, "Learning to Collocate Neural Modules for Image Captioning," in ICCV, 2019.

31. S. Herdade, A. Kappeler, K. Boakye, and J. Soares, "Image Captioning: Transforming Objects into Words," in NeurIPS, 2019

32. L. Huang, W. Wang, J. Chen, and X.-Y. Wei, "Attention on Attention for Image Captioning," in ICCV, 2019.

33. W. Liu, S. Chen, L. Guo, X. Zhu, and J. Liu, "CPTR: Full Transformer Network for Image Captioning," arXiv preprint arXiv:2101.10804, 2021.